\documentclass{article}
\usepackage{spconf,amsmath,amsfonts,graphicx}

\usepackage{amssymb}
\usepackage{url}

\title{ROI-based Deep Image Compression with Swin Transformers}
\name{Binglin Li$ ^{\star} $,  Jie Liang$ ^{\star} $, Haisheng Fu$ ^{\dagger} $, Jingning Han$ ^{\ddagger} $} 
\address{$ ^{\star} $ School of Engineering Science, Simon
Fraser University, Canada, \{binglinl, jiel\}@sfu.ca \\
$ ^{\dagger} $School of Microelectronics, Xi’an Jiaotong University, China, fhs4118005070@stu.xjtu.edu.cn \\
$ ^{\ddagger} $Google LLC, Mountain View, CA, USA, jingning@google.com}

\begin{document}
%
\maketitle
\begin{abstract}
Encoding the Region Of Interest (ROI) with better quality than the background has many applications including video conferencing systems, video surveillance and object-oriented vision tasks. In this paper, we propose a ROI-based image compression framework with Swin transformers as main building blocks for the autoencoder network.
The binary ROI mask is integrated into different layers of the network to provide spatial information guidance. 
Based on the ROI mask, we can control the relative importance of the ROI and non-ROI by modifying the corresponding Lagrange multiplier $ \lambda $ for different regions. 
Experimental results show our model achieves higher ROI PSNR than other methods and modest average PSNR for human evaluation. When tested on models pre-trained with original images, it has superior object detection and instance segmentation performance on the COCO validation dataset.
\end{abstract}
\section{Introduction}
\label{sec:intro}
Region of interest (ROI)-based image compression encodes the ROI with better quality than the background which has many applications such as video conferencing systems and video surveillance. More bits allocated for the ROI means more distinguishable human faces and behaviors than the surroundings, therefore it improves conference experience and tracking performance. It can also be utilized in medical images where the regions with organs have higher research interest to physicians and the background is less important~\cite{tahoces2008image}.

The classic image coding standard JPEG2000 \cite{jpeg2000} supports the ROI coding by scaling up more important coefficients in ROI and placing them in higher bitplanes. In~\cite{han2006image, han2008roi}, the authors generate the saliency map as a probability density of detected salient points and use JPEG2000 for coding with the saliency map. Nevertheless, it suffers from severe discontinuity at the boundary between the ROI and non-ROI.
Some learning-based ROI image compression works are proposed later.
For example, a weighted Multi-Scale Structural Similarity Index Measure (MS-SSIM) loss is applied as the quality control of priority parts and non-priority parts in~\cite{Akutsu_2019_CVPR_Workshops, xia2020object}. In~\cite{cai2019end, ma2021variable}, a combination of Mean Squared Error (MSE) loss in ROI and MS-SSIM loss for other necessary regions is used. 
In~\cite{patelsaliency}, a sum of MSE loss and deep perceptual loss with different weights for salient and non-salient regions is adopted.
In these works, the ROI bit allocation is achieved by applying the ROI mask to the feature representation before quantization and/or combining with the ROI-aware loss during training.


Our approach is motivated by~\cite{song2021variable}, where a quality map with continuous values is embedded to different layers in the network for variable-rate image compression, and a set of $ \lambda $s are mapped and trained in a single model. Although it is mentioned that the task-aware image compression can be realized at test time, it has not been clearly explored for ROI-based image compression. In our work, we focus on improving the compression performance in ROI while maintaining good overall visual quality for human evaluation. Based on the ROI mask, we can control the relative importance of the ROI and non-ROI by modifying the corresponding $ \lambda $ for different regions.

Swin transformers~\cite{liu2021swin} are used for the autoencoder network because they can extract the long-range dependencies in the image. Vision transformers have been applied in deep image compression~\cite{lu2021transformer, li2022variable} for non-ROI methods. The transformers in~\cite{lu2021transformer} act like non-local modules. In~\cite{li2022variable}, the compression performance is limited due to the patch partition. We build the network with Swin transformer blocks with consideration of computational cost. 

The superior ROI compression performance can benefit computer vision tasks compared with non-ROI methods. Our work is different from the task-driven deep image compression~\cite{patelsaliency, li2022region} and high-fidelity image compression~\cite{mentzer2020high, ma2021variable}. In these problems, the loss function can be modified with a task-based loss or Generative Adversarial Network (GAN) loss to improve the computer vision performance or the perceptual similarity to original images. 
This is usually compromised with the decrease of PSNR, which is commonly used in image compression benchmarking. We train the compression network with MSE as distortion loss and show computer vision results tested on models pre-trained with original images.

Our contributions are summarized as follows:\\
\indent 1) A ROI-based image compression network with Swin transformers~\cite{liu2021swin} is established. Different from previous ROI-based methods for image compression, the ROI mask is incorporated through the entire network to give the spatial information about salient regions. Experiments verify that our method achieves better compression results for ROI as well as modest overall reconstruction quality by setting an appropriate mapping $ \lambda $ for the background. We also explore both of the ground truth and predicted ROI masks. It shows that the accuracy of the ROI prediction is critical for results.

2) We use Swin transformers as main building blocks.  Different behaviors of attention maps for the transformer blocks in the encoder and decoder networks are observed, which suggest that the encoder may need larger receptive filed than the decoder for autoencoder-based network in learned image compression.

3) Experiments show that our ROI-based image compression model achieves superior object detection and instance segmentation results compared with non-ROI methods when tested on the COCO validation dataset~\cite{lin2014microsoft}. This demonstrates that the ROI-based image compression can benefit object-oriented vision tasks even without designing more complex loss functions.

\begin{figure}[t] 
\centerline{\includegraphics[width=85mm]{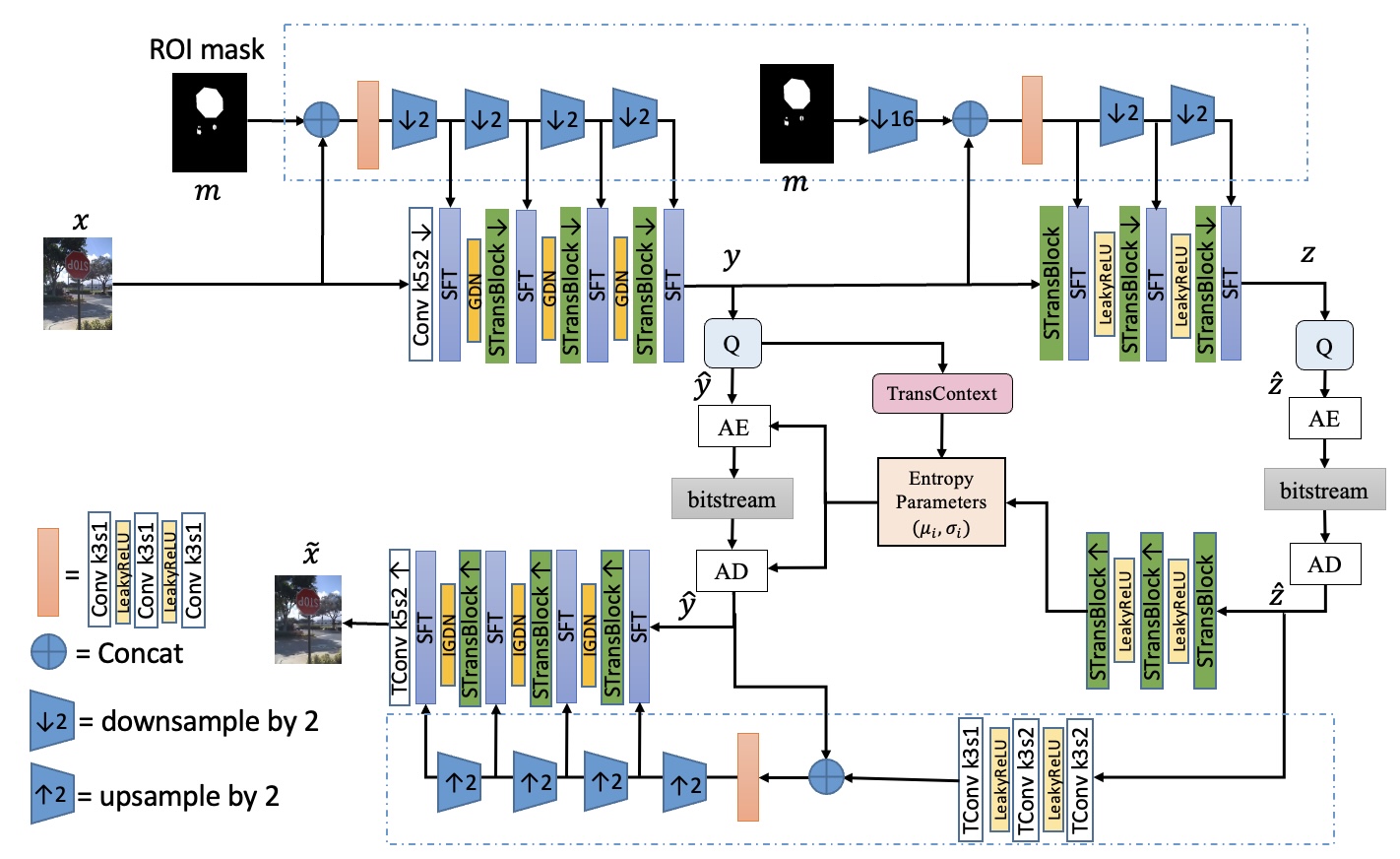}}
\caption{The proposed ROI-based deep image compression architecture. STransblock represents the Swin transformer block~\cite{liu2021swin} modified for upsampling and downsampling~\cite{li2022variable}.}
\label{fig:overview}
\end{figure}


\section{The Proposed Framework}
\label{sec:approach}
We consider ROI-based deep image compression that the ROI is allocated with more bits than the non-ROI. The overall compression network is given in Fig.~\ref{fig:overview}.  
It consists of an autoencoder network path for image compression and a ROI information conveying path as shown in dashed box.
The input ROI mask can be ground truth or estimated from other networks. 
Different from previous methods for image compression, the ROI mask is embedded into different layers in the main and hyper-prior networks which allows to learn discriminative feature representations for different regions during training.

The autoencoder network follows the hierarchical architecture in~\cite{minnen2018joint}. The main encoder and decoder are comprised of Swin transformer blocks~\cite{liu2021swin}, Generalized Divisive Normalization (GDN)~\cite{balle2015density} layers and Spatially-adaptive Feature Transform (SFT) blocks~\cite{song2021variable}. The GDN layers are generally applied in deep image compression models~\cite{balle2015density, minnen2018joint} to realize local divisive normalization for density modeling. As the computational complexity of transformers is a quadratic function of the sequence length, we use one convolutional layer with $ 5\times 5 $ kernels and stride size of 2 at the first layer of main encoder to downsample and the last layer of main decoder to upsample the input tensor by a factor of 2 respectively. 
The context model TransContext is the same as used in~\cite{li2022variable}.

\subsection{Swin Transformer Blocks}
A Swin transformer block contains a multi-head self-attention (MHSA) network and a point-wise feedforward network (FFN). The consecutive blocks are computed by Window-based MHSA (W-MHSA) and Shifted Window-based MHSA (SW-MHSA)~\cite{liu2021swin} as
\begin{equation}
\begin{aligned}
& \hat{z}^{l} = \text{W-MHSA}(\text{LN}(z^{l-1}))+z^{l-1}, \\
& z^{l} = \text{FFN}(\text{LN}(\hat{z}^{l}))+\hat{z}^{l}, \\
& \hat{z}^{l+1}  = \text{SW-MHSA}(\text{LN}(z^{l}))+z^{l}, \\
& z^{l+1} = \text{FFN}(\text{LN}(\hat{z}^{l+1}))+\hat{z}^{l+1}, \\
\end{aligned}
\end{equation}
where $ \hat{z}^{l} $ and $ z^{l} $are output of the W-MHSA/SW-MHSA network and the FFN network for the block $ l $ respectively. 

The SW-MHSA is introduced for cross-window connection which is proved beneficial for vision tasks~\cite{liu2021swin}.
We use window size of $ 8\times 8 $ for the main encoder and decoder, and $ 4 \times 4 $ for the hyper-encoder and hyper-decoder. We modify the Swin transformer blocks for downsampling with patch partitioning and upsampling with patch merging~\cite{li2022variable}. 

\begin{figure*}[!t]
\begin{tabular}{cccc}
   \includegraphics[width=0.21\linewidth,height=32mm]{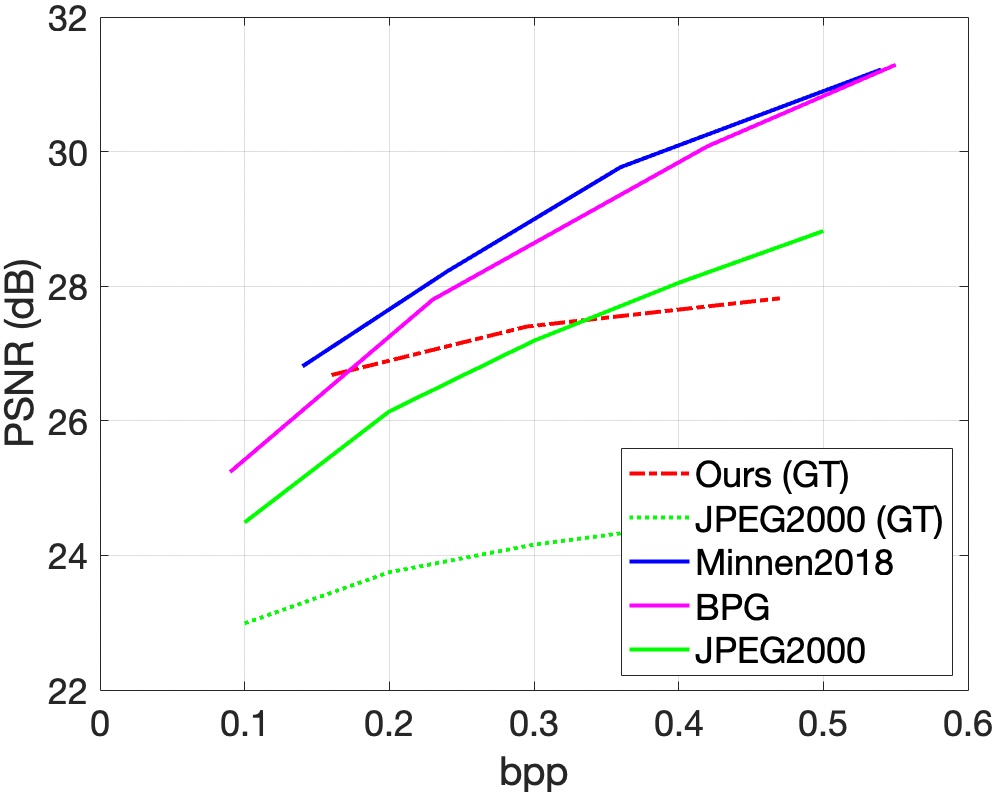}&
   \includegraphics[width=0.21\linewidth,height=32mm]{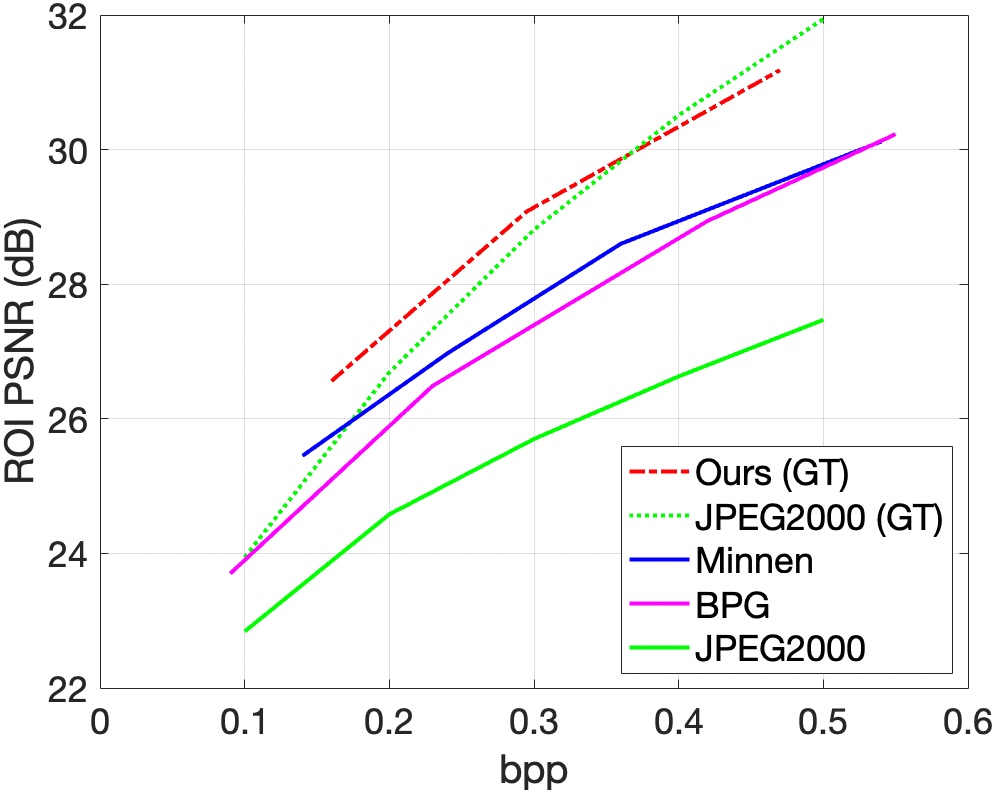} &
        \includegraphics[width=0.21\linewidth,height=32mm]{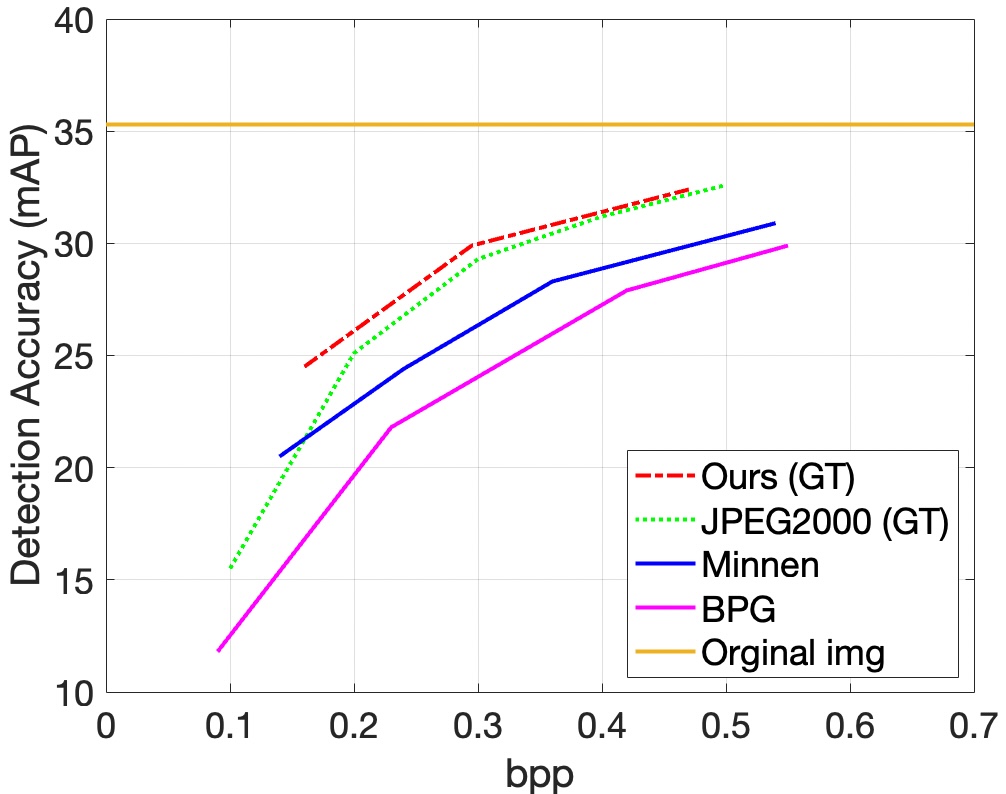}&
   \includegraphics[width=0.21\linewidth,height=32mm]{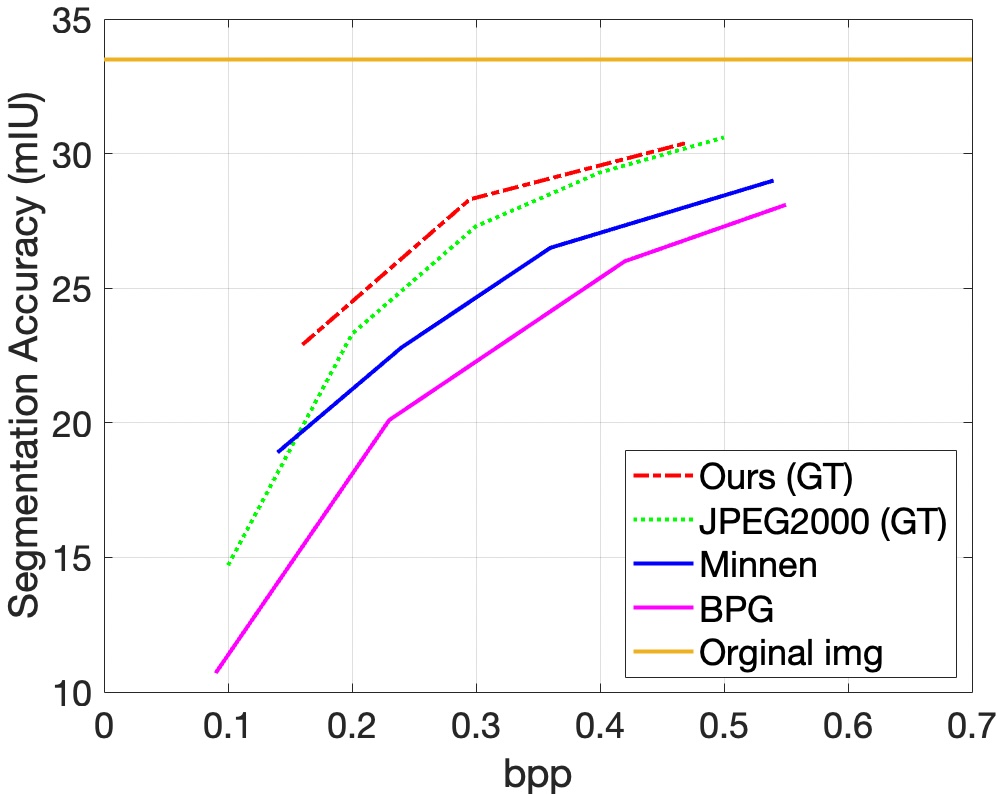} \\
\end{tabular}
   \caption{From left to right: (a) PSNR, (b) ROI PSNR, (c) object detection, (d) instance segmentation on COCO validation dataset~\cite{lin2014microsoft}.}
\label{fig:psnr_results}
\end{figure*}

\subsection{Spatially-adaptive Feature Transform blocks}
Instead of pre-defining four kinds of quality map as in~\cite{song2021variable} during training, we use the ground truth or predicted ROI mask which is integrated to intermediate feature tensors by the SFT blocks as shown in Fig.~\ref{fig:overview}. 
The ROI mask is first concatenated with the input image and then input to some convolutional layers to get fused information. We simplify the architecture in~\cite{song2021variable} and only use average pooling operations to match the spatial size for different layers. The total number of parameters for our proposed framework is around 15.8M. The SFT blocks integrate the ROI spatial information all through the network to learn discriminative feature representations for different regions during training.

SFT blocks learn a parameter pair $ (\gamma, \beta) $ based on given prior condition $ \Psi $ which is related to the ROI map $ m $ in our case for spatial-wise feature modulation. 
Based on this, an affine transformation is applied spatially to the intermediate feature maps with the parameter pair~\cite{song2021variable} as
\begin{equation}
\text{SFT}(F, \Psi) = \gamma \odot F +\beta
\end{equation}
where $ F $ denotes an intermediate feature map. 
As the decoder may not have the ROI mask, the input for SFT module is obtained from the hyper-latent $ \hat{z} $ as it also contains the spatial information from the hyper-encoder~\cite{song2021variable}.

\section{Experiment}
\label{sec:experiment}

\subsection{Dataset and Training Details}
We train the models on the COCO 2017 training set~\cite{lin2014microsoft} which contains 118k images. The 5k validation images are used to validate the ROI PSNR performance. 
The learning rate is set to 0.0001 and batch size is set to 8. We run on one12GB  TITAN X GPU card with Adam optimizer. The training lasts 50 epochs.

The loss function is $ L = \lambda D + R $ during training, where $ D $ is the MSE between the original image $ x $ and reconstructed image $ \tilde{x} $. $ R $ is the sum of bit rates for latent $ \hat{y} $ and hyper-latent $ \hat{z} $ as $ R =E[-log_{2} p_{\hat{y}|\hat{z}} (\hat{y}|\hat{z})]+ E[-log_{2}p_{\hat{z}}(\hat{z})] $.
The tradeoff $ \lambda $ is related to the binary ROI mask $ m\in \lbrace 0,1 \rbrace $. The ROI mask $ m $ is projected to the corresponding $ \lambda $ to allocate the different bits for ROI and non-ROI by $ \lambda_{i} = \alpha e^{\omega  m_{i}} \times 255^{2}$.  The ratio of bits assigned to ROI and non-ROI can be adjusted by $ \alpha $ and $ \omega $. $ \alpha $ is a parameter to set $ \lambda $ for background where $ m_{i} $ is close to 0. Note that some bits are kept at the non-ROI to maintain the average compression performance and visual experience. 
We set $ \alpha=0.001 $ during experiment. $ \omega $ is a parameter that can be used to modify the amount of bits assigned to the ROI. We set $ \omega=\lbrace 3, 5, 6.5 \rbrace $ to obtain results for different bit rates. 


\begin{figure}[!t]
\begin{center}
\includegraphics[width=\linewidth,height=45mm]{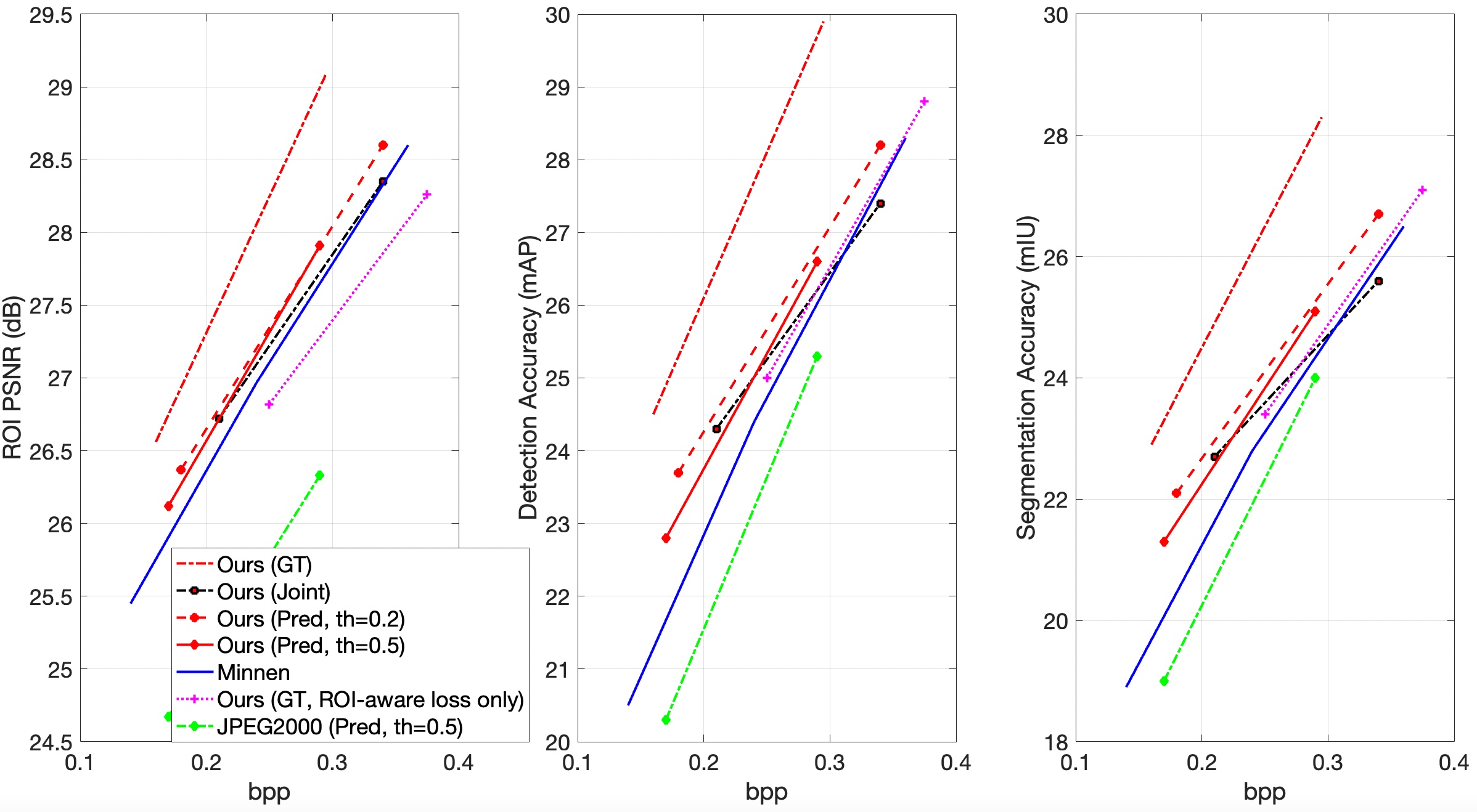}
\end{center}
   \caption{Influence of ROI Mask. From left to right: (a) ROI PSNR, (b) object detection, (c) instance segmentation.}
\label{fig:blation_test}
\end{figure}

\begin{figure*}[!t]
\begin{tabular}{ccccccc}
   \includegraphics[width=0.11\linewidth,height=15mm]{fig/real_0003.png}&
   \includegraphics[width=0.11\linewidth,height=15mm]{fig/jp2k.png} &
   \includegraphics[width=0.11\linewidth,height=15mm]{fig/bpg44.png} &
   \includegraphics[width=0.11\linewidth,height=15mm]{fig/mbt_q1.png}&
        \includegraphics[width=0.11\linewidth,height=15mm]{fig/jp2k_roi.png}&
   \includegraphics[width=0.11\linewidth,height=15mm]{fig/ours_pred.png} &
     \includegraphics[width=0.11\linewidth,height=15mm]{fig/ours_gt.png}\\   
   
      \includegraphics[width=0.11\linewidth]{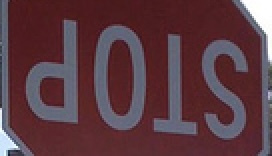}&
   \includegraphics[width=0.11\linewidth]{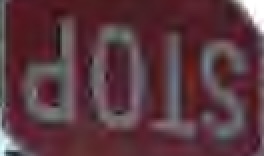} &
   \includegraphics[width=0.11\linewidth]{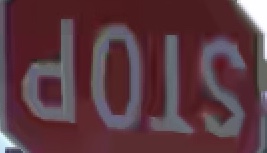} &
   \includegraphics[width=0.11\linewidth]{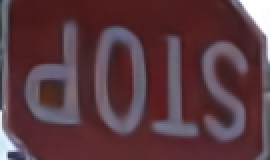}&
        \includegraphics[width=0.11\linewidth]{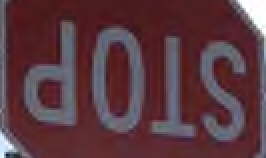}&
   \includegraphics[width=0.11\linewidth]{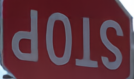} &
     \includegraphics[width=0.11\linewidth]{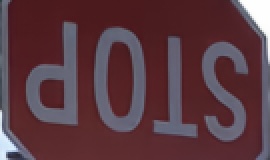}\\		
    ROI PSNR/bpp & 23.47/0.20 & 27.34/0.22 & 28.38/0.19 & 29.15/0.20 & 30.64/0.20 & 34.48/0.21 \\    
    Original image & JPEG2000 & BPG &  Minnen~\cite{minnen2018joint} & JPEG2000 (GT) &  Ours (Pred) & Ours (GT) \\
   \end{tabular}
\begin{tabular}{ccccccc}
   \includegraphics[width=0.11\linewidth,height=12mm]{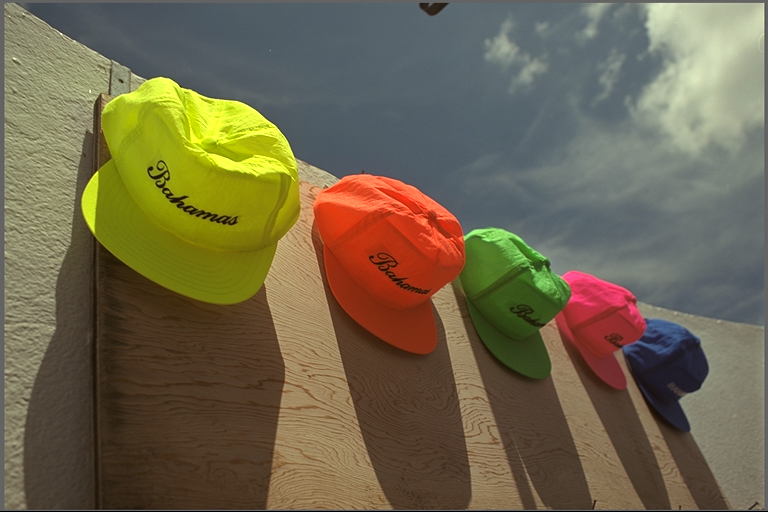}&
   \includegraphics[width=0.11\linewidth,height=12mm]{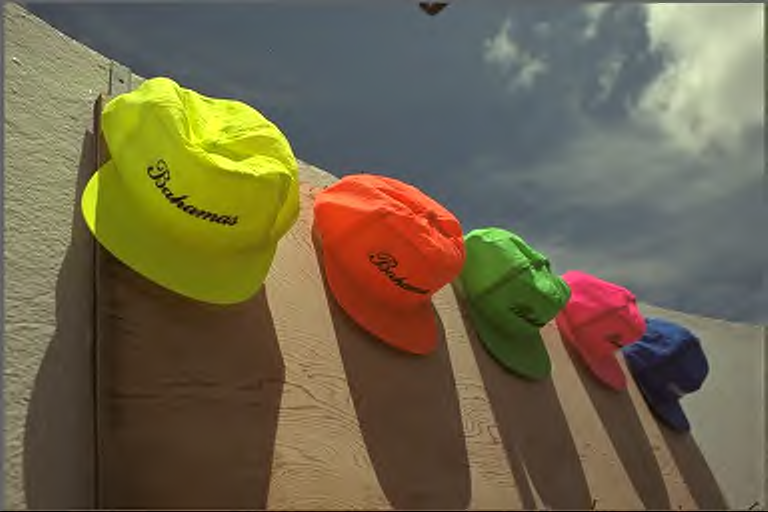} &
   \includegraphics[width=0.11\linewidth,height=12mm]{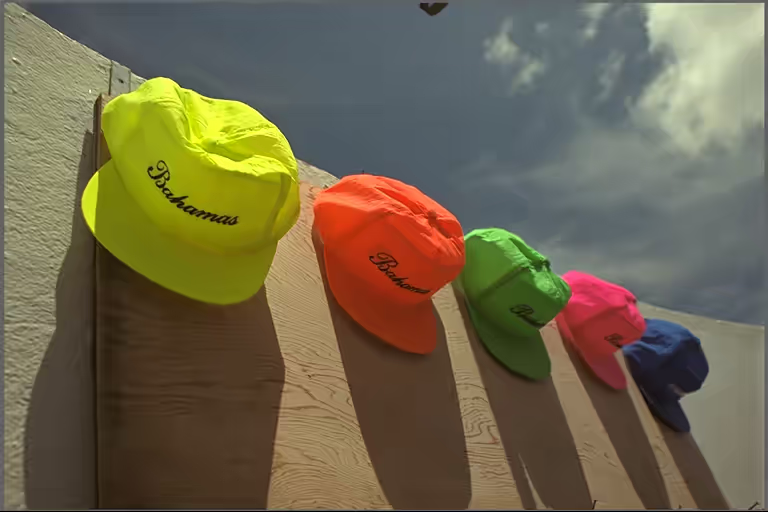} &
    \includegraphics[width=0.11\linewidth,height=12mm]{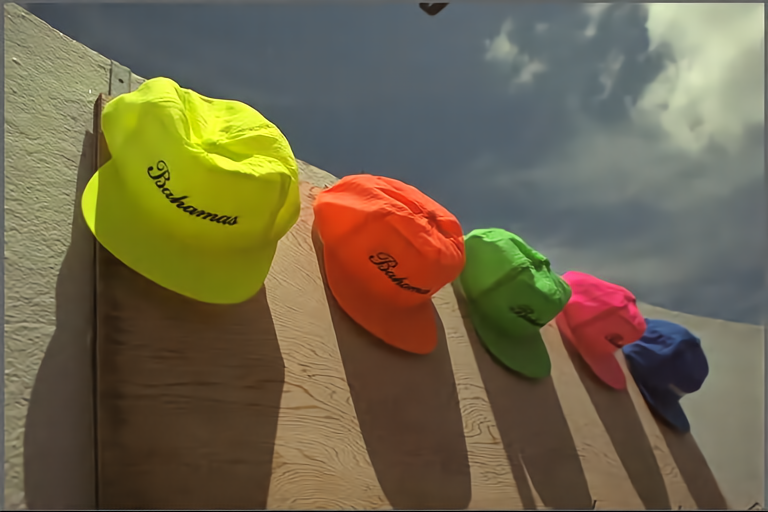}&  
   \includegraphics[width=0.11\linewidth,height=12mm]{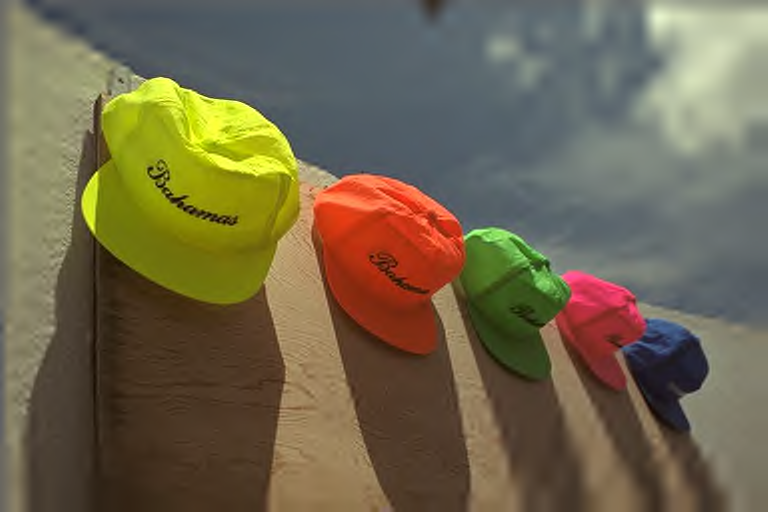} &
   \includegraphics[width=0.11\linewidth,height=12mm]{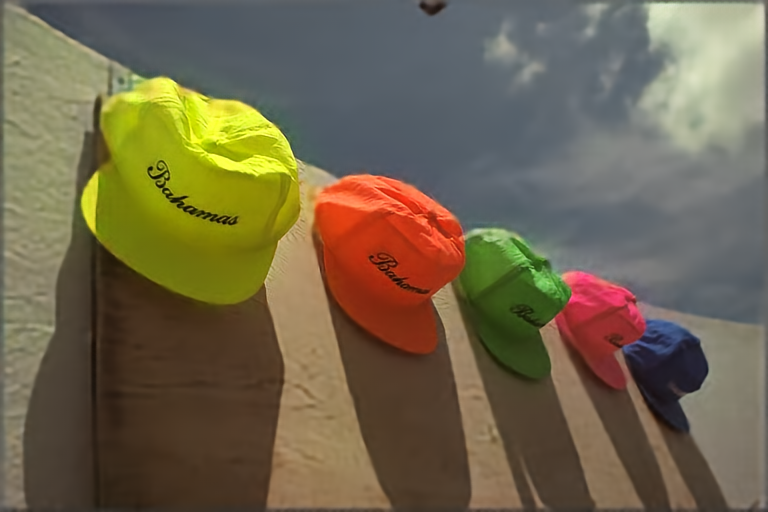}&
   \includegraphics[width=0.11\linewidth,height=12mm]{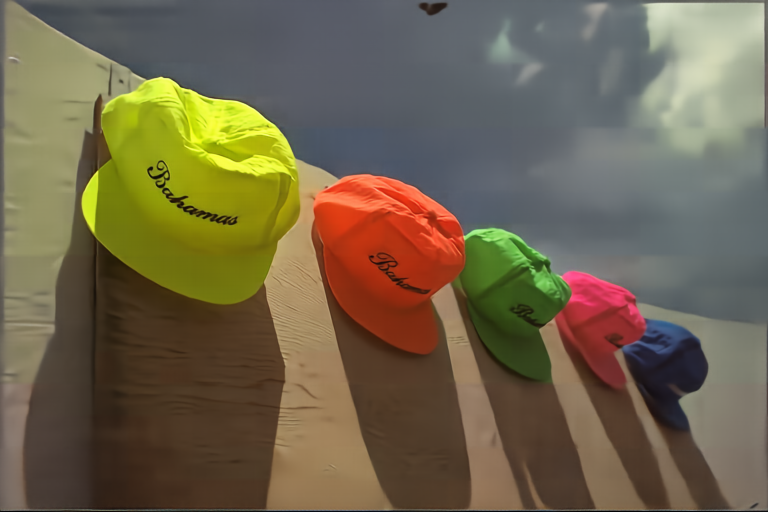}\\   
   
    \includegraphics[width=0.11\linewidth,,height=13mm]{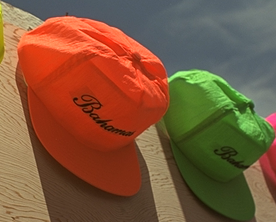}&
    \includegraphics[width=0.11\linewidth,height=13mm]{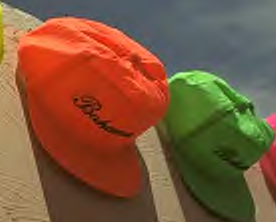}&
   \includegraphics[width=0.11\linewidth,height=13mm]{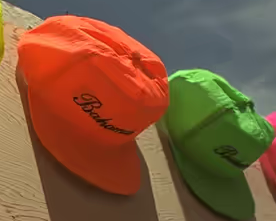} &
   \includegraphics[width=0.11\linewidth,height=13mm]{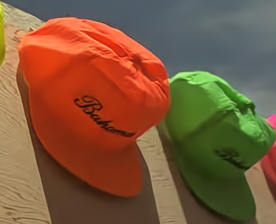}&
    \includegraphics[width=0.11\linewidth,height=13mm]{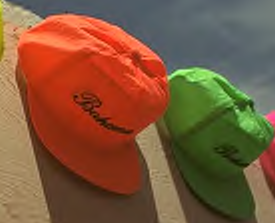} &
    \includegraphics[width=0.11\linewidth,height=13mm]{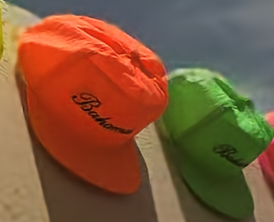}&
    \includegraphics[width=0.11\linewidth,height=13mm]{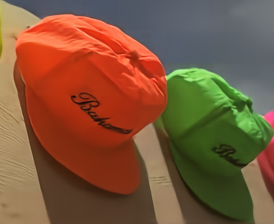}\\
PSNR/bpp   & 31.95/0.20 & 35.15/0.22 & 35.78/0.21 & 28.84/0.20 & 30.18/0.19 & 31.15/0.17 \\
Original image & JPEG2000 & BPG &  Minnen~\cite{minnen2018joint}  &  JPEG2000 (Pred) & Cai~\cite{cai2019end} & Ours (Pred)\\
\end{tabular}
\caption{Comparisons on images with different methods.}
\label{fig:example_results}
\end{figure*}

\subsection{Results on COCO Validation Dataset}
We show average PSNR and ROI PSNR over bit per pixel (bpp) on the COCO validation dataset in Fig.~\ref{fig:psnr_results} (a) and (b). Ours (GT) and JPEG2000 (GT)~\footnote{https://kakadusoftware.com} are obtained from the ground truth ROI Masks. We also compare with non-ROI methods BPG~\footnote{http://bellard.org/bpg/}, \textit{Minnen}~\cite{minnen2018joint} and JPEG2000. The results for \textit{Minnen} are obtained from the pre-trained models in~\cite{begaint2020compressai}. 

For average PSNR in (a), ours (GT) performs much better than JPEG2000 (GT) across a range of bit rates and comparable to BPG at 0.16bpp. Compared to BPG and \textit{Minnen} at higher bit rates, the average PSNR from ours (GT) is worse due to effect of shifting bits from non-ROI to ROI.
For ROI PSNR in (b), ours (GT) shows superior performance than all other methods at low bit rates. JPEG2000 (GT) has better ROI PSNR at higher bit rates. Note that we keep some backgrounds for human evaluation by setting $ \alpha=0.001 $. We could reduce $ \alpha $ to assign more bits in ROI to further improve ROI performance by tolerating some decrease of non-ROI quality.
Fig.~\ref{fig:example_results} displays example results obtained from different methods at the first row. Ours (GT) has more clear reconstruction for the stop sign than other methods and better overall visual quality for the whole image than JPEG2000 (GT).
Ours (Pred) is obtained from the predicted ROI mask which will be described next.

We evaluate the reconstructed images from different methods by testing on the models pre-trained with original images for object detection using Faster-RCNN~\cite{ren2015faster} and instance segmentation using Mask-RCNN~\cite{he2017mask}. The original result is 35.3\% for object detection and 33.5\% for instance segmentation.
In Fig.~\ref{fig:psnr_results} (c) and (d), ours (GT) outperforms BPG and \textit{Minnen}. Compared with JPEG2000 (GT), the detection and segmentation performance also have some increase at low bit rates. 




\textbf{Influence of ROI Mask}
We also experiment to adopt the HRNet~\cite{wang2020deep} to predict the ROI masks.
Fig.~\ref{fig:blation_test} shows the results with ground truth (GT), an offline predicted (Pred) with different thresholds (th=0.2 or th=0.5) and a joint training predicted (Joint) ROI masks for our method.  
Results using GT ROI have a significant gain over those using predicted ROI.
Overall, our method shows better performance than JPEG2000 ROI coding for both scenarios. 
For ours (Pred, th=0.2) performs slightly better than th=0.5 due to less false negatives.
Ours (Joint) achieves similar result as ours (pred) at 0.21bpp but worse at high bit rate. It is probably because the higher $ \lambda $ leads to less penalty for ROI loss and lower ROI accuracy is obtained.
The results from ROI-aware loss are obtained by applying the true ROI mask to distortion loss and the layers in dashed box in Fig.~\ref{fig:overview} are deleted. The degraded performance demonstrates the necessity of incorporating spatial ROI information to intermediate layers.
It is trained from the same weight ratio for ROI and non-ROI as ours (GT). 

\subsection{Results on Kodak Dataset}
As ground truth ROI masks are not provided for Kodak dataset~\footnote{http://r0k.us/graphics/kodak/}, we show example comparisons at the second row in Fig.~\ref{fig:example_results}. 
The predicted mask covers five hats and some shadow areas.
Our (Pred) has clearer textures for the green hat than other methods and more continuity at ROI boundaries than JPEG2000 (Pred) and Cai~\cite{cai2019end} with less bit rate, although the average PSNR is degraded than non-ROI methods.

\begin{figure}[t!]
\begin{center}
   
\includegraphics[width=0.9\linewidth,height=22mm]{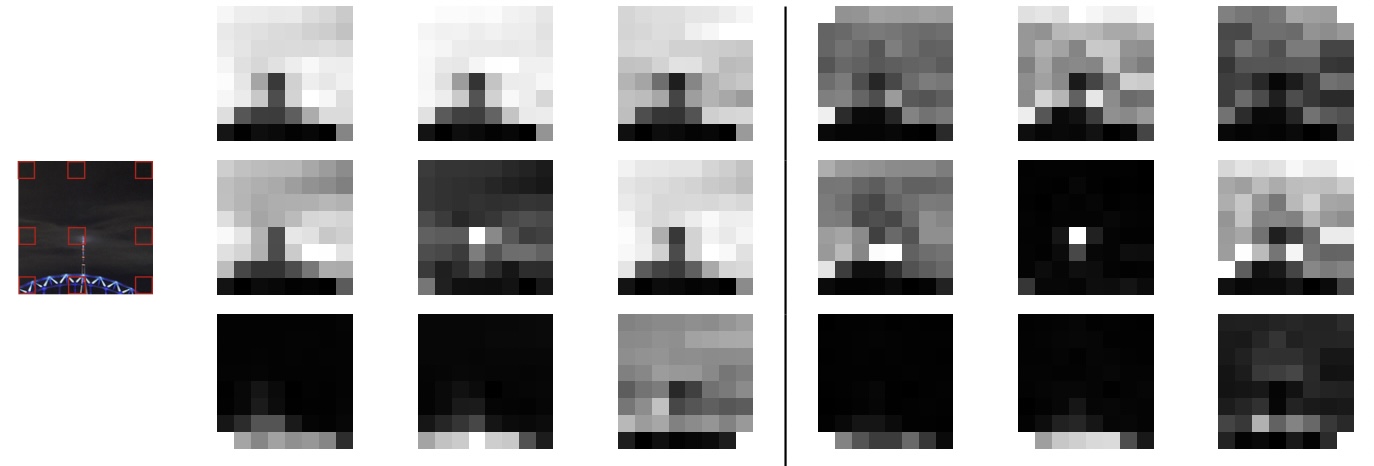} 
\end{center}
   \caption{Examples of attention maps. 
  }
\label{fig:example_attention}
\end{figure}

\subsection{Attention Maps in Swin Transformers} 
We show examples of attention maps corresponding to nine receptive fields in Fig.~\ref{fig:example_attention}. For points in smooth area (top), the neighboring points have higher attention scores. For points in patterned area (bottom), it attends to the points at the bottom. In addition, more high response points in the encoder (left columns) are observed than the points in the decoder (right columns), which suggest that the encoder may need more neighboring and complex information than the decoder.

\section{Conclusion}
\label{sec:conclusion}
We propose a ROI-based deep image compression framework with Swin transformers. 
The binary ROI mask is incorporated into different layers of the network to provide spatial information guidance. We experiment on ground truth ROI and predicted ROI to show the influence of ROI accuracy for the results. 
Experimental results show our model achieves better ROI PSNR while keeping modest average PSNR, thus contributing to higher object detection and instance segmentation performance. 



\end{document}